\documentclass[letterpaper, 10 pt, conference]{ieeeconf}  % Comment this line out if you need a4paper

\IEEEoverridecommandlockouts                              % This command is only needed if 
\usepackage{microtype}
\usepackage{graphicx}
\usepackage{epigraph}
\usepackage[cmex10]{amsmath}
\DeclareMathOperator*{\argmax}{arg\,max}

\usepackage{amsfonts}
\usepackage{setspace}
\usepackage{layout}
\usepackage[subfigure]{tocloft}  % Set TOC options.
\usepackage{lipsum}
\usepackage{pgfplots}
\usepackage{pgfplotstable}
\usepackage{tikz}
\pgfplotsset{compat=1.9}
\usetikzlibrary{positioning}
\usetikzlibrary{backgrounds}
\usetikzlibrary{calc}
\usepgfplotslibrary{colorbrewer}  % provided by local file.
\usepgfplotslibrary{statistics}

\usepackage[format=hang]{subfig}  % Do not include caption package in here.
\usepackage{xcolor}  % Get extra colors.
\usepackage{algorithm}
\usepackage{algpseudocode} % algorithmicx
\usepackage{ctable}
\usepackage{multirow}

\usepackage{xspace} % macros with trailing spaces
\usepackage{siunitx}
\renewcommand{\baselinestretch}{0.936}

\usepackage{float}
\newfloat{algorithm}{tbp}{lop}

\title{\LARGE \bf
A POMDP-based hierarchical planning framework for manipulation under pose uncertainty
}

\author{Muhammad Suhail Saleem, Rishi Veerapaneni, and Maxim Likhachev% <-this % stops a space
\thanks{All authors are with the Robotics Institute, Carnegie Mellon University, Pittsburgh, PA 15213, USA. {\small e-mail: \tt \{msaleem2, rveerapa, mlikhach\}@andrew.cmu.edu}
}
}

\begin{document}

\maketitle
\thispagestyle{empty}
\pagestyle{empty}

%%%%%%%%%%%%%%%%%%%%%%%%%%%%%%%%%%%%%%%%%%%%%%%%%%%%%%%%%%%%%%%%%%%%%%%%%%%%%%%%
\begin{abstract}

Robots often face challenges in domestic environments where visual feedback is ineffective, such as retrieving objects obstructed by occlusions or finding a light switch in the dark. In these cases, utilizing contacts to localize the target object can be effective. We propose an online planning framework using binary contact signals for manipulation tasks with pose uncertainty, formulated as a Partially Observable Markov Decision Process (POMDP). Naively representing the belief as a particle set makes planning infeasible due to the large uncertainties in domestic settings, as identifying the best sequence of actions requires rolling out thousands of actions across millions of particles, taking significant compute time. To address this, we propose a hierarchical belief representation. Initially, we represent the uncertainty coarsely in a 3D volumetric space. Policies that refine uncertainty in this space are computed and executed, and once uncertainty is sufficiently reduced, the problem is translated back into the particle space for further refinement before task completion. We utilize a closed-loop planning and execution framework with a heuristic-search-based anytime solver that computes partial policies within a limited time budget. The performance of the framework is demonstrated both in real world and in simulation on the high-precision task of inserting a plug into a port using a UR10e manipulator, resolving positional uncertainties up to 50 centimeters and angular uncertainties close to $2\pi$. Experimental results highlight the framework's effectiveness, achieving a 93\% success rate in the real world and over 50\% improvement in solution quality compared to greedy baselines, significantly accelerating planning and enabling real-time solutions for complex problems.

% Robots often face challenges in domestic environments where visual feedback is ineffective, such as retrieving objects obstructed by occlusions or finding a light switch in the dark. In these cases, utilizing contact to localize the target object can be effective. We propose an online planning framework using binary contact signals for manipulation tasks with pose uncertainty, formulated as a Partially Observable Markov Decision Process (POMDP). Naively representing the belief as a particle set makes planning infeasible due to the large uncertainties in domestic settings, as identifying the best sequence of actions requires rolling out thousands of actions across millions of particles, taking significant compute time. To address this, we propose a hierarchical belief representation, initially representing uncertainty coarsely in a 3D volumetric space. Policies that refine uncertainty in this space are computed and executed, and once uncertainty is sufficiently reduced, the problem is translated back into the particle space for further refinement before task completion. We utilize a closed-loop planning and execution framework with a heuristic-search-based anytime solver that computes partial policies within a limited time budget, balancing exploration and exploitation. Demonstrations in real-world and simulated tasks, such as plug insertion, show a 93\% success rate and a 50\% improvement in solution quality compared to greedy baselines, significantly accelerating planning and enabling real-time solutions for complex problems.

\end{abstract}

%%%%%%%%%%%%%%%%%%%%%%%%%%%%%%%%%%%%%%%%%%%%%%%%%%%%%%%%%%%%%%%%%%%%%%%%%%%%%%%%
\section{INTRODUCTION}

\begin{figure}
\centering
\includegraphics[width=\linewidth]{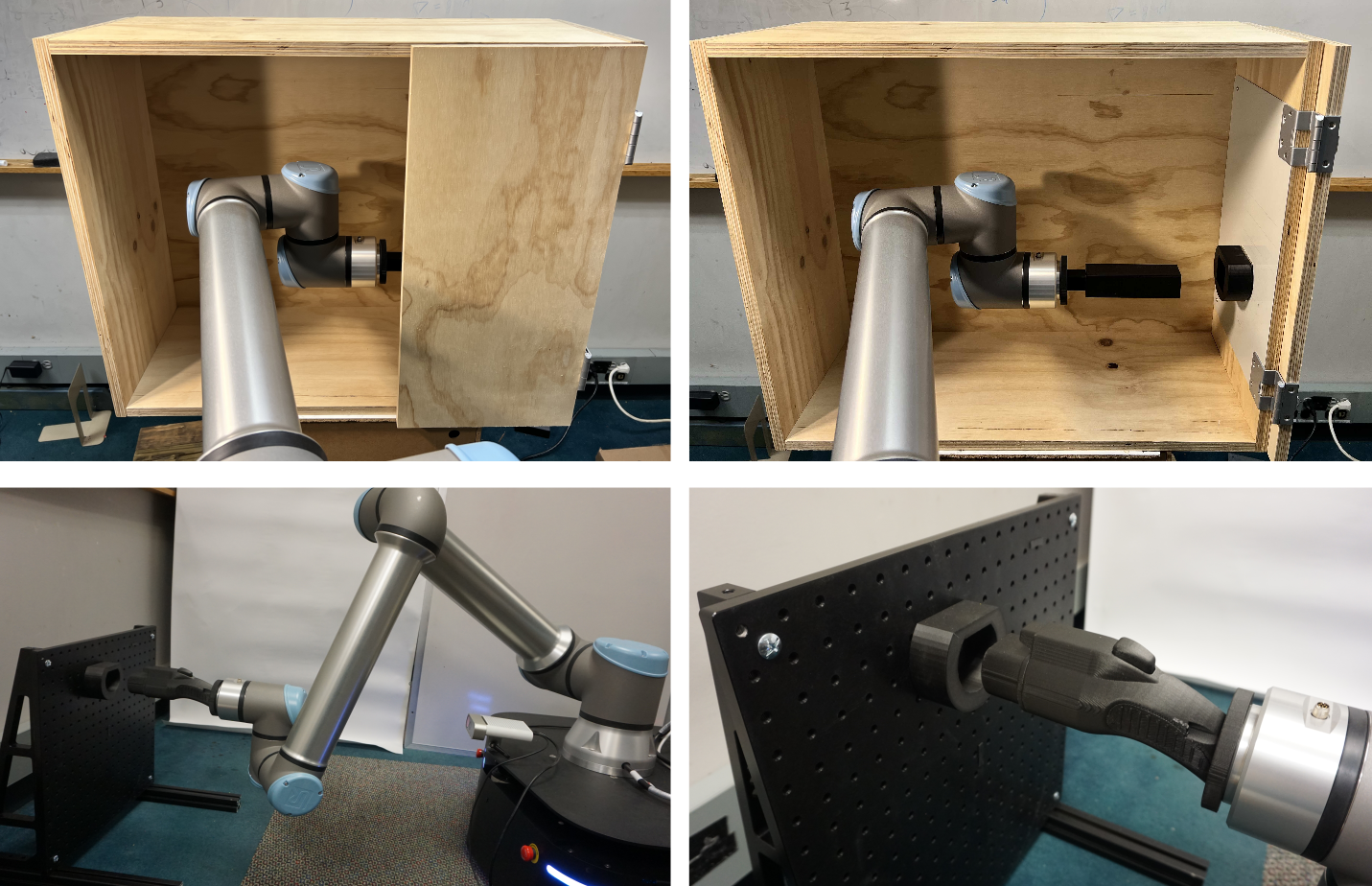} 
\caption{\footnotesize Experimental setup for plug insertion using a UR10e manipulator. Top: The port is located inside a shelf and out of view of the robot (uncertainty along $y, z, roll$). Bottom: Movable base set up (uncertainty along $x, y, z, yaw$).} \label{fig:shelf_setup}
\vspace{-1em}
\end{figure}

Robots frequently encounter challenges in domestic settings where relying solely on visual feedback proves inadequate or inefficient. For instance, consider Fig. \ref{fig:shelf_setup} where the robot is tasked with locating an object within a shelf and occlusions obstruct the robot's view. Similar visibility issues arise when a robot tries to locate the light switch in a dark room or in situations like furniture assembly, where one arm is dedicated to stabilizing larger components while the other is retrieving smaller components like fasteners. A common strategy that emerges in all of these scenarios is leveraging contact feedback to localize the objects of interest. 
% In these scenarios, the robot should contact the object of interest at different locations enabling it to localize the target object without relying on its visual capabilities. 
This mirrors the intuitive approach humans take in similar situations.

% Robots frequently encounter challenges in domestic settings where relying solely on visual feedback proves inadequate or inefficient. For instance, a robot may need to locate an object within a shelf where occlusions obstruct its view, similar to the situation in Fig. \ref{fig:shelf_setup}. Similar visibility issues arise when a robot tries to locate a switch in a dark room or when assembling furniture, where one arm is stabilizing larger components while the other retrieves smaller parts. In all these scenarios, leveraging contact feedback to localize objects is an effective strategy. The robot should contact the object of interest at different locations, enabling it to localize the target object without relying on visual capabilities. This approach mirrors the intuitive methods humans use in similar situations.

In this manuscript, we propose a planning framework for less structured domestic settings that enables a robot to use binary contact signals to localize a target object sufficiently enough to complete the task at hand. The problem of planning for active information gathering to complete a task can be formulated as a Partially Observable Markov Decision Process (POMDP) \cite{intro_POMDP_2} \cite{intro_sondik1971optimal} \cite{intro_POMDP_1}. POMDPs are a general framework for planning in partially observable environments. However, they are notoriously difficult to solve, and for most real-world problems due to the large state space and planning horizon, solving a POMDP is computationally intractable \cite{intro_POMDP_2} \cite{intro_zhou}. 

% The active localization problems discussed in this manuscript are complex, and solving them exactly and in real-time is infeasible, calling for the development of a smart planning solution. Further, the variety of problems in domestic settings also requires the planning framework to handle large magnitudes of uncertainty effectively. If we take the example highlighted in Fig. \ref{ch4_fig: shelf_setup} since the robot can not directly observe the object of interest, it has to rely on the prior that the port typically resides on the right wall of the shelf. This implies that the robot now has to deal with positional uncertainties of the order of tens of centimeters and angular uncertainty closer to $2\pi$. Similarly, when a robot enters a dark room and is trying to locate the light switch which it knows is located somewhere on the wall right next to the door, it has to deal with uncertainties in position ranging closer to a meter. Hence, the order of pose uncertainties that need to be resolved is large. 

The active localization problems discussed in this manuscript are complex, and solving them exactly and in real-time is infeasible, necessitating a smart planning solution. Additionally, the variety of problems in domestic settings requires the planning framework to handle large uncertainties effectively. For example, in the shelf scenario in Fig. \ref{fig:shelf_setup}, since the robot cannot directly observe the object of interest, it has to rely on its prior knowledge that the target object typically resides on the right wall of the shelf. This implies resolving positional uncertainties of tens of centimeters and angular uncertainties close to $2\pi$. Similarly, locating a light switch in a dark room involves positional uncertainties up to a meter.

% The large magnitude of pose uncertainty manifests itself in the planning problem in two forms. First, it significantly escalates the complexity of the planning problem. Due to the large amounts of uncertainty, the number of branches in the POMDP is substantially large, making the planning problem challenging \cite{intro_zhou} \cite{ch4_complexity} \cite{ch4_complexity_2}. Second, a POMDP enables one to systematically reason about the different outcomes upon executing a sequence of actions and pick the sequence that maximizes an objective. An important element in this problem is computing the belief and how the belief can transition when an action is executed. Given the magnitude of the uncertainty in this setting, representing the belief naively as a particle set would mean that we have millions of particles. Computing the outcome of an action for each of these particles individually is a computationally expensive affair. Given that multiple sequences of actions have to be rolled out while solving the POMDP, it is impractical to represent the belief as a particle set. This calls for a smarter representation of belief. 

The large magnitude of pose uncertainty presents two key challenges. First, it significantly escalates the complexity of the planning problem by leading to a substantial increase in the number of branches in the POMDP, making the planning process challenging \cite{intro_zhou} \cite{ch4_complexity} \cite{ch4_complexity_2}. Second, a POMDP enables one to systematically reason about different outcomes from executing a sequence of actions to select the sequence that maximizes an objective. A critical aspect here is computing the belief and its transitions when an action is executed. Given the significant uncertainty, naively representing the belief as a particle set results in millions of particles. For each particle representing a target object pose, computing the expected observation for an action requires compute-heavy mesh-to-mesh collision checks. Identifying the best sequence of actions involves rolling out thousands of actions across millions of particles, which is computationally intensive. This necessitates a smarter representation of belief. To address this, we develop a planning framework that utilizes the following:

\begin{enumerate}
    \item \textbf{A hierarchical representation of uncertainty:} Uncertainty is initially represented coarsely in a 3D volumetric space, capturing the volume potentially occupied by the object of interest. Contact and no-contact observations from different actions provide information that shrinks this potentially occupied volume. Once the uncertainty is sufficiently reduced in this space, we translate the problem back into the particle space to refine finer uncertainties and complete the task. 
    % This hierarchical approach allows for quick reasoning under large uncertainties and the computation of intelligent policies online.    
    % We initially represent and reduce the uncertainty in a coarse 3D volumetric space. After computing and executing policies in this space, we translate the problem back into the particle space to refine finer uncertainties and complete the task.

    \item \textbf{A greedy closed-loop planning and execution framework:} 
    In each iteration of planning, the planner has a limited time budget (1 second) and a heuristic-search-based anytime solver is used to compute partial policies. To maximize performance within a limited time budget, the planner efficiently utilizes its resources by i) Focusing on situations that are most likely to occur, and ii) Combining an admissible and an inadmissible heuristic to guide the search (achieving a balance between exploration and exploitation).
\end{enumerate}

The performance of the overall framework is demonstrated both in the real world and in simulation on the task of inserting a plug into a port using a UR10e manipulator in the presence of different magnitudes of pose uncertainties (3D, 4D, and 5D). The port is positioned outside the robot's view, requiring the robot to resolve positional uncertainties of the order of tens of centimeters and angular uncertainties close to $2\pi$ to complete the insertion task. Experimental results highlight the robustness of the framework, achieving a 93\% success rate. Performance analyses, presented in Section \ref{section: Results}, show that the proposed hierarchical representation of belief significantly accelerates planning, making it feasible to solve such complex problems online. Additionally, the planning framework improves solution quality by over 50\% compared to greedy baselines. 

\section{RELATED WORK}
Over the years, various vision-based techniques have been developed for pose estimation of objects for manipulation tasks such as high-precision plug insertions (a task that we demonstrate our framework on) \cite{kragic2002survey}\cite{zou2023object}\cite{xiang2017posecnn}. Visual servoing is a widely used approach for such tasks, utilizing real-time image data from an in-hand camera to compute errors—either directly in image space or by estimating pose of the insertion slot—and generating control commands to minimize these errors \cite{ch3_IBVS1}\cite{ch3_PBVS1}\cite{ch3_HBVS1}\cite{ch3_VS_learning}. However, in this work, we are focused on scenarios where visual feedback is infeasible or ineffective, and instead, the robot must rely solely on contact feedback.

Many prior works that incorporate tactile feedback focus on developing particle filters and Bayesian estimation methods to compute object poses that best explain the sequence of tactile observations made \cite{ch3_global_localization_via_touch}\cite{ch3_TBL_for_parts}\cite{ch3_saundPHD}\cite{petrovskaya2006bayesian}. However, our work focuses more on active localization using tactile feedback, i.e. reasoning about the sequence of actions to take to quickly localize the object of interest. Previous works tackling this question have utilized a set of particles to represent the object pose uncertainty \cite{ch3_TBL}\cite{saleem2023preprocessing}. Directly formalizing and solving the POMDP by utilizing this representation is infeasible in situations with large uncertainty, prevalent in unstructured environments, as millions of particles may be needed. Other particle filtering variants such as \cite{koval2015pose} require compute heavy operations like manifold sampling, making them unsuitable for planning. 
In this work, we address this challenge by proposing a hierarchical representation of uncertainty and defining the POMDP using this representation. 

Our hierarchical representation is independent of the specific planner used to solve the POMDP, allowing for efficient action sequence reasoning. Prior approaches have used greedy methods to approximately solve the POMDP, often employing a myopic framework that interleaves planning and execution \cite{ch3_TBL}\cite{ch3_TBL_for_parts}\cite{ch4_yamauchi}. In each iteration of touch-based localization (TBL) \cite{ch3_TBL}, actions are sampled, and the robot executes the one that maximizes an information gain metric. This process is repeated until uncertainty is sufficiently reduced. Similarly, frontier search \cite{ch4_yamauchi} selects the nearest information-gain action for execution. However, these myopic strategies, which focus only on the next best action, often compromise solution quality. Performance comparisons against these baselines presented in Section \ref{section: Results} highlight this issue. 

The transition models for the proposed representation of uncertainty is closely tied to \cite{ch4_saund2020motion}\cite{ch3_saundPHD}\cite{ch4_saund2023blindfolded}. However, their objective is to utilize contact observations to model the unknown environment they operate in and efficiently move from one location to another. On the planning front, RTDP-Bel \cite{ch3_RTDP-Bel} forms the basis of our framework. Further,  \cite{ch4_satia_lave}\cite{ch4_online_POMDP} discuss techniques to focus efforts on the most likely outcomes/beliefs when planning online under time constraints, an idea that we adapt and utilize in our framework. Finally, our work is closely related to \cite{saleem2023preprocessing}, which addresses the problem of active information gathering using contact-based manipulation for semi-structured industrial settings. Here, the limited uncertainty and structure enables them to preprocess a database of solution policies that can be queried online.

\section{PROBLEM FORMULATION}
Given a robot manipulator, let $\mathcal{Q}$ represent its state space and $\mathcal{A}$ the discrete action space. The observation space $\mathcal{Z}$ comprises of
\begin{enumerate}
    \item The robot's state (encoder readings), and
    \item A binary flag indicating the occurrence of contacts/collisions (F/T sensor readings or joint torques)
\end{enumerate}

Let $T$ represent a target object of interest. $V$ represents a prior on the pose of the target object defined as a volume in the 3D workspace within which the target object is guaranteed to be. The goal of the planning problem is to compute a robot policy $\pi$ that utilizes contacts as observations to reduce uncertainty about the pose of $T$ to below a task-dependent threshold $\epsilon$ and complete the task while minimizing the expected distance traveled.

The planning problem described is addressed using a hierarchical representation of pose uncertainty. A detailed description of our hierarchical formulation and the exact POMDP solved can be found in the section below. Unlike typical robotics problems, the stochasticity in transitions and observations arises from uncertainty in the model of the environment, specifically the object pose. We assume perfect dynamics and observations, due to the high-quality manipulators and the simple but reliable observation space (contacts) 
\footnote{
% To validate the absence of noise in contact observations, we sampled 100 random actions. Given a known pose of the target object we observed that in all 100 cases, the expected and observed collision readings aligned perfectly.
We validated perfect contact observations by sampling 100 actions and confirming that all collision readings aligned with the expected observations.}
. The environment is assumed to be static, with robot actions not altering the state of $T$. We also assume a realizable setting, i.e, $T$ is guaranteed to be within the initial hypothesis volume $V$.

% The assumptions we make can be concisely listed as follows: 
% Succinctly, we make the following assumptions:
% \begin{itemize}
%     \item[\textbf{A1}] Realizable setting, i.e., the object of interest is guaranteed to be within the initial hypothesis volume $V$.
%     \item[\textbf{A2}] Deterministic transitions and perfect observations given the environment state, i.e., object pose. 
%     \item[\textbf{A3}] Known geometric model of the object of interest (used to compute expected observations).
%     \item[\textbf{A4}] Static environment, i.e., the object of interest remains static upon interaction.
% \end{itemize}

\section{METHODOLOGY}

\begin{figure*}
\centering
  \includegraphics[width=0.85\textwidth]{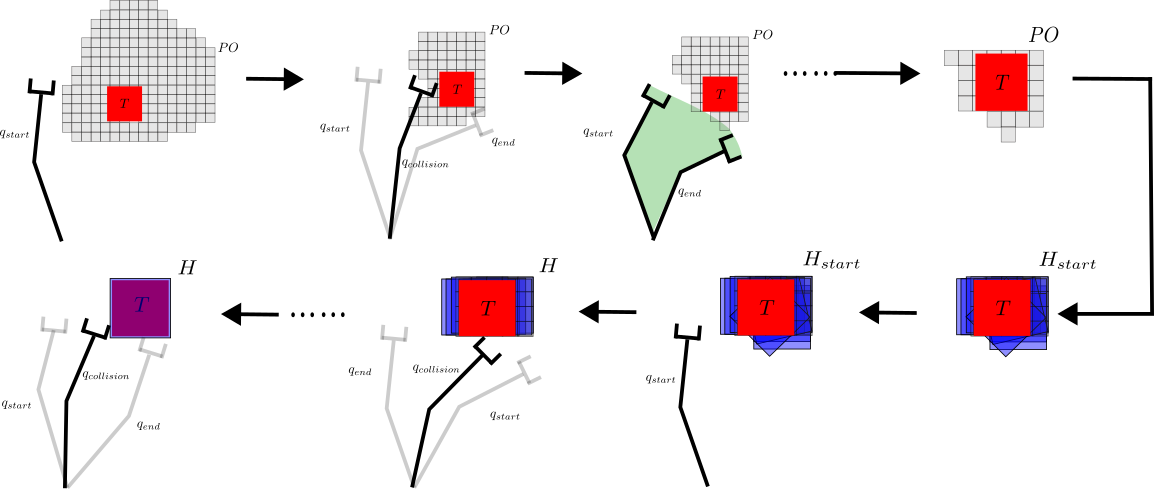}
  \caption{
\footnotesize The first phase of the framework involves reasoning about pose uncertainty in the volumetric space, using contact observations from actions to reduce the potentially occupied volume. Once uncertainty is reduced to a manageable level, we transition to a finer representation in the particle space by mapping the reduced volumetric uncertainty to a smaller set of feasible poses. We then execute policies that will help reduce uncertainty in this space sufficient enough to complete the task.
}\label{fig: overall framework}
\end{figure*}

% Given a particle representing a target object pose, computing the expected observation for an action requires compute-heavy mesh-to-mesh collision checks. When uncertainties are large, identifying the best sequence of actions involves rolling out thousands of actions across millions of particles, which is computationally intensive. To reduce this burden, we propose a hierarchical uncertainty representation. Essentially, 
The planning framework is split into two phases (Fig. \ref{fig: overall framework}). 

\textbf{Phase 1 (Volumetric representation): }We start by representing uncertainty in a 3D volumetric space, which captures the potential volume occupied by the target object. As actions are executed, observations (e.g., collisions) help refine this space. For example, if an action results in no collision, we can infer that the target object does not occupy the swept volume, reducing the possible occupied space. This approach allows for efficient reasoning under large uncertainties and helps identify actions that minimize the potentially occupied volume. A customized heuristic search-based POMDP solver is used to compute a policy that reduces the uncertainty in the volumetric space below a predefined threshold.

\textbf{Phase 2 (Particle representation): }Once the uncertainty is reduced to a manageable level, we transition to a finer representation in the particle space. We map the reduced volumetric uncertainty into a smaller set of feasible target poses (particles). A new POMDP is defined to represent uncertainty transitions in this particle space, and the same heuristic solver is employed to compute a policy that further reduces uncertainty and completes the task.

By hierarchically representing uncertainty, our approach reduces computational complexity, effectively shrinking the search space and allowing a manageable transition from volumetric to particle-based reasoning. The following subsections detail the POMDP formulation in both phases and the customized solver used.

\subsection{Hierarchical Belief Representation}

\begin{figure}
\centering
  \includegraphics[width=0.4 \textwidth]{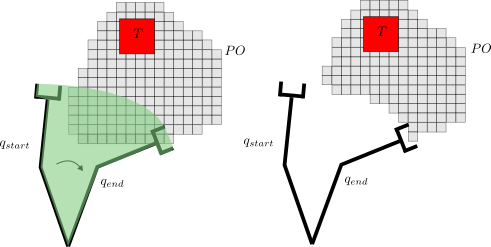}
  \caption{\footnotesize If an action results in no contact, the target object is guaranteed to not occupy any part of the swept volume.}\label{fig: sweeping explanation (no collision)}
\end{figure}

\begin{figure}
\centering
  \includegraphics[width=0.4 \textwidth]{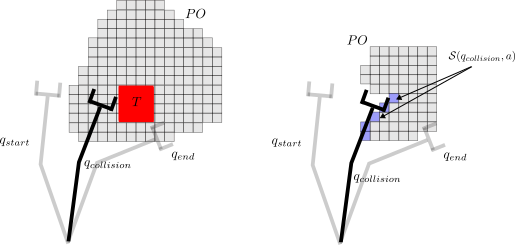}
  \caption{\footnotesize If an action $a$ results in contact at configuration $q_{collision}$, the target object is guaranteed to not be farther than $\mathcal{D}_{max}$ from $\mathcal{S}(q_{collision}, a)$.}\label{fig: sweeping explanation (collision)}
  \vspace{-1em}
\end{figure}

\subsubsection{Volumetric representation}
The volumetric representation of uncertainty is structured as a 3D binary occupancy grid, where each voxel has a value of 0 or 1. A voxel with a value of 0 indicates that the object of interest is guaranteed to not occupy any part of it, while a voxel with a value of 1 represents the possibility of occupancy. Of the total set of voxels, we denote the set of possibly occupied voxels as $PO$, where $|PO|$ represents the number of such voxels.

We start by discretizing the initial volume $V$ into a 3D grid, assigning 1 to all cells. As 
$PO$ is not directly observable, collision observations from executed actions help reduce 
$|PO|$. The goal of the planning problem at this phase is to find a robot policy that minimizes the expected travel distance while reducing $|PO|$ below a predetermined threshold $\delta$.

The system's state $s$ comprises the robot's state, the environment state, and the set of previously encountered collisions represented as $s = (q, PO, \mathcal{C})$. Where $\mathcal{C} = \{(q_1, a_1), (q_2, a_2) ... \}$ and each tuple $(q_i, a_i) \in \mathcal{C}$ corresponds to the robot having previously observed a collision at configuration $q_i$ while executing $a_i$. The action space $\mathcal{A}$ consists of end-effector movements. The observation space $\mathcal{Z}$ includes binary collision readings and the robot state 
$\{\text{(no) collision}, q \in \mathcal{Q}\}$. A robot action $a$ starting from $q_1$ and ending at $q_n$ can be approximated by a discrete set of configurations $\{q_1, q_2 ... q_n\}$. Such an action can produce $n+1$ possible observations: collisions at any intermediate configuration $z_i = \{\text{collision},q_i\}$ or no collision upon reaching $q_n$, denoted $z_{n+1} = \{\text{no collision}, q_n\}$. 

The successor state $s'$ from executing action $a$ from state $s = \{q, PO, \mathcal{C}\}$ and receiving observation $z$ is as follows:
% observation $z$ was made is as follows:

\begin{enumerate}
    \item If $z_{n+1}$ is observed (no collision), $s' = \{q_n, PO_{n+1}, \mathcal{C}_{n+1}\}$, where:
        \begin{equation} \label{PO_no_collision_update}
            PO_{n+1} = PO \, \backslash \, \mathcal{V}(q_1, q_n); \,\,\, \mathcal{C}_{n+1} = \mathcal{C}
        \end{equation}
    % \item If $z_{n+1}$ is observed (no collision), $s' = \{q_n, PO \, \backslash \, \mathcal{V}(q_1, q_n), \mathcal{C}\}$ where
        Here $\mathcal{V}(q_1, q_n)$ represents the set of voxels contained in the volume swept by the robot moving from $q_1$ to $q_n$ (Fig. \ref{fig: sweeping explanation (no collision)}).
    \item If $z_{i}$ is observed (collision at $q_i$), $s' = \{q_i, PO_{i}, \mathcal{C}_i\}$, where:
        \begin{equation} \label{PO_collision_update}
            PO_{i} = PO \, \backslash \left( \mathcal{V}(q_1, q_i) \, \bigcup \, \mathcal{U}(q_i, a) \right); \,\, \mathcal{C}_{i} = \mathcal{C} \cup (q_i, a)
        \end{equation}
        If a collision was observed at $q_i$ when executing $a$, let $\mathcal{S}(q_i, a)$ represent the set of voxels (on the surface of the robot) out of which at least one is guaranteed to be occupied. If $\mathcal{D}_{max}$ represents the maximum distance between any two points on the object of interest, all cells farther than $\mathcal{D}_{max}$ from $\mathcal{S}(q_i, a)$ are guaranteed to be unoccupied (Fig. \ref{fig: sweeping explanation (collision)}). This set is defined by $\mathcal{U}(q_i, a)$ and can be eliminated.
\end{enumerate}

Hence, given a state $s = \{q, PO, \mathcal{C}\}$, action $a$, and the set of possible observations $Z$, it is possible to compute the set of successor states. To compute the transition function $\mathcal{T}(s', s, a)$, we determine the likelihood of each observation. Let $N$ be the number of voxels occupied by the target object $T$. Then, assuming independence amongst the observations, we define the likelihood of collision and no collision at configuration $q_i$ as,
\begin{equation*} 
\begin{split}
 \phi(\{&\text{collision}, q_i\}| PO, a, \mathcal{C}) \propto \\ & \phi(\{\text{collision}, q_i\}| PO, a) \hspace{-0.3cm} \prod_{(q_j, a_j) \in \mathcal{C}} \hspace{-0.3cm} \phi(\{\text{collision}, q_i\}| (q_j, a_j), a) \\
&\text{Where, }\\
& \phi(\{\text{collision}, q_i\} | PO, a) = \min(1, \frac{|\mathcal{S}(q_i, a) \cap PO|}{|PO| - N}) 
% \\
% & \phi(\{\text{no collision}, q_i\} | PO, a) = 1 - \phi(\{\text{collision}, q_i\} | PO, a)
\end{split}
\end{equation*}
\begin{equation} \label{eqn: PO likelihood}
\phi(\{\text{no collision}, q_i\} | PO, a) = 1 - \phi(\{\text{collision}, q_i\} | PO, a)
\end{equation}
\vspace{0.2cm}
\begin{equation*} 
\begin{split}
\phi(\{\text{collision}, q_i\}| (q_1, &a_1), a) = 
    (\frac{|\mathcal{S}(q_i, a) \cap \mathcal{S}(q_1, a_1)| + \epsilon}{|\mathcal{S}(q_1, a_1)| + \epsilon}) \times \\ & (1 - \frac{\max(0, dist(\mathcal{S}(q_i, a), \mathcal{S}(q_1, a_1)))}{\mathcal{D}_{max}})
\end{split}
\end{equation*}
\begin{equation} \label{eqn: C likelihood}
\begin{split}
\phi(\{\text{no collision}, q_i\}| &(q_1, a_1), a) = \\
                                   &1 - \phi(\{\text{collision}, q_i\}| (q_1, a_1), a)
\end{split}
\end{equation}

The likelihood is essentially composed of two modules. First, Eqn. \ref{eqn: PO likelihood} defines the likelihood of a configuration being in collision as proportional to the volume of $PO$ it intersects. Next, Eqn. \ref{eqn: C likelihood} consists of two components. The first component defines that the likelihood of collision increases as the volume of overlap between $\mathcal{S}(q_i, a)$ and $\mathcal{S}(q_j, a_j)$ (where $(q_j, a_j) \in \mathcal{C}$) increases, with a collision observation being guaranteed if $\mathcal{S}(q_i, a) \subseteq \mathcal{S}(q_j, a_j)$. The second component represents the decreasing likelihood of collision as the distance between the voxel sets increases, with the likelihood being zero if the distance exceeds $\mathcal{D}_{max}$.

The likelihood of the different observations needs to account for the fact that the observations are sequential. If $z_i$ is observed, no collision must have occurred at earlier configurations $\{q_1, q_2 ... q_{i-1}\}$. Thus:

\begin{equation} \label{ch4_eqn: PO collision likelihood}
\begin{split}
    \phi(z_i | a, s = \{q, PO\}) &= \phi(\{\text{collision}, q_i\} | PO, a) \, \times \\ & \prod_{k = [1, i-1]} \phi(\{\text{no collision}, q_k\} | PO, a)  
\end{split}
\end{equation}

% \begin{equation} \label{ch4_eqn: PO no collision likelihood}
%     \phi(z_{n+1} | a, s) = \prod_{k = [1, n]} \phi(\{\text{no collision}, q_k\} | PO, a)  
% \end{equation}

The likelihood of $z_{n+1}$ can also be similarly computed. These likelihoods provide the observation model $P(z| s, a)$, which, together with the state transitions defined by $PO_{n+1}$ and $PO_i$, completes the transition model $T(s'|s, a)$. This formalizes the problem in volumetric space, allowing a belief MDP solver to compute a policy that minimizes the expected distance while reducing $|PO|$ below $\delta$.

\subsubsection{Particle Representation}
After reducing the set of potentially occupied voxels, we switch the uncertainty representation to particle space. This involves discretizing the space of possible poses\footnote{The resolution of discretization is dependent on the tolerance of the task.} and evaluating each pose to determine if the object occupies any portion of the workspace outside the potentially occupied voxels set. If a pose results in the object occupying space outside this voxel set, it is considered infeasible as it conflicts with previous observations. Conversely, if the pose is feasible, it is included in the set of valid hypothesis poses. 

Now that the set of hypothesis poses $H_{start} = \{h_1, h_2, ... h_n\}$ has been created, we define the POMDP in the particle space similar to \cite{saleem2023preprocessing}. The state of the system in this case contains the robot state and the set of particles $s = \{q, H\}$. An action $a \in \mathcal{A}$ approximated by a sequence of $n$ discrete configurations $\{q_1, q_2 ... q_n\}$ as previously discussed can result in $n+1$ observations. Hence, if an observation $z_i$ is made (which corresponds to collision at $q_i$), the system transitions to a new state $s' = \{q_i, \Hat{H}\}$, where $\Hat{H}$ corresponds to all particles (i.e, object poses) for which executing action $a$ from $q$ would result in a collision at $q_i$. Given a particle $h_i$, the expected contact observation can be computed by collision checking the discretized approximation of the action with the target object located at pose $h_i$. If the probability of each particle in the set is uniform, then the probability of observing $z_i$ is equal to $\frac{|\Hat{H}|}{|H|}$. 

This completes the definition of the POMDP in the particle space (more details can be found in \cite{saleem2023preprocessing}). Now, a belief MDP solver can be used to compute a policy from the start state $s_{start} =\{q_{start}, H_{start}\}$ to a state $s_{goal} = \{q_{goal}, H_{goal}\}$ where $s_{goal}$ satisfies the goal criteria for all the particles in $H_{goal}$. In the case of the plugin problem, this corresponds to a robot pose $s_{goal}$ such that the charger has been successfully plugged into the port for all poses in $H_{goal}$.

% This completes the definition of the POMDP in the particle space. A belief MDP solver can be used to compute a policy that drives the system from a start state $s_{start} = \{q_{start}, H_{start}\}$ to a goal state $s_{goal} = \{q_{goal}, H_{goal}\}$, where $q_{goal}$ satisfies the goal criteria for all particles in $H_{goal}$. For example, in the case of the plugin problem, the goal state corresponds to a robot configuration where the charger has been successfully plugged into the port for all possible object poses in $H_{goal}$. 

% For both the particle and volumetric representation, we utilize a heuristic search-based online POMDP solver, the details of which we elucidate in the following subsection \ref{ch4_subsection: Planning Framework}.  

\subsection{Planning Framework} \label{ch4_subsection: Planning Framework}

% In the previous subsection, we explored the representation of belief and the formulation of the planning problem as a POMDP. Here, we detail the planning framework employed for solving the formulated POMDPs. 
Here, we detail the planning framework employed for solving the POMDPs formulated in the previous subsection. We note the overall framework discussed earlier is agnostic to the specific planner used. Our empirical analysis indicate that the planner described below yielded the best results.

Computing complete (bounded sub-)optimal policies for the problems discussed in this manuscript can take extensive compute time (of the order of hours). Since the planner must solve these problems online, we use a closed-loop planning and execution framework with a 1-second time budget for each iteration. In each iteration, we use a modified version of RTDP-Bel (Alg. \ref{Alg: Modified RTDP-Bel}), a heuristic search-based anytime solver to reason over a limited time horizon and compute a partial policy, which is then executed in the real world. If the system reaches a belief state without a precomputed action, the planner is invoked again. This process repeats until the task is completed.
To maximize performance, we modify RTDP-Bel in two ways.

\begin{algorithm}
\begin{small}
\caption{\textsc{Modified RTDP-Bel}}
\label{Alg: Modified RTDP-Bel}
\begin{algorithmic}[1]
\While {\textsc{Not Converged}}
\State $b = b_{start}$; $depth = 0$ \label{Alg2:Start}
\While {$b \notin \mathcal{G}$ and $depth < horizon$}
    \For {type $\in$ $\{admissible,\, inadmissible\}$} \label{alg_line: ad, inad begin}
        \State \textbf{Evaluate} the values of executing each action $a \in \mathcal{A}$ \\ \hspace{1.3cm} from belief state $b$ as: 
        % \State 
        \begin{equation*} \vspace{-0.15cm}
        Q_{type}(b, a) = \mathcal{C}(b, a) + \sum_{z \in \mathcal{Z}} P(z | b, a) V_{type}(b_a^z)
        \end{equation*}
        \Comment{\parbox[t]{.65\linewidth}{When value not initialized, use  \vspace{-0.15cm}
        \begin{equation*}
            V_{type}(b_a^z) = heur_{type}(b_a^z)
        \end{equation*}
        }}
        \State \label{Alg2: Update Value}\textbf{Update} value of belief state  \vspace{-0.15cm}
        \begin{equation*}  \vspace{-0.15cm}
                V_{type}(b) = \min_{a \in \mathcal{A}} Q_{type}(b, a)
        \end{equation*}
    \EndFor
\If{time elapsed $< \epsilon \text{ } *$ time budget} \Comment{$\epsilon \in [0, 1]$}
\State \textbf{Select} action $a_{best}$ that minimizes $Q_{ad}(b, a)$
\Else 
\State \textbf{Select} action $a_{best}$ that minimizes $Q_{inad}(b, a)$
\EndIf \label{alg_line: ad, inad end}
\State \textbf{Pick} most likely $b'$, i.e., $\argmax_{b' \in \mathcal{B}} P(b' | b, a_{best})$ \label{alg_line: most likely b}
\State \textbf{Set} $b := b'$ and $depth := depth + 1$
\EndWhile
\EndWhile
\end{algorithmic} 
\end{small}
\end{algorithm}

\subsubsection{Compute actions only for the most probable outcomes} 

Inspired by existing strategies in literature \cite{ch4_online_POMDP} \cite{ch4_satia_lave}, we optimize the use of limited resources by focusing computational efforts on the most likely scenarios. Specifically, we modify the planner to only compute actions for the most probable beliefs. 
In each iteration within an episode in RTDP-Bel, the best action from a belief state is computed based on the current value estimates of the successor beliefs from the state. While we consider all possible outcomes when determining the best action, the subsequent belief state that we explore is restricted to the most probable successor. In the subsequent iteration the best action for the most probable successor belief is computed (Line \ref{alg_line: most likely b}). If, during the execution of the partial policy, a less likely successor belief is reached, execution is halted, and the planner is invoked again. This approach ensures the planner’s limited time is spent reasoning about the most likely situations, enhancing overall effectiveness.

\subsubsection{Combining admissible and inadmissible heuristics}
The effectiveness of search-based planners relies heavily on the heuristics employed. Heuristics guide the search to explore relevant portions of the belief space, ideally reducing computational costs. An admissible heuristic, which underestimates the optimal cost ($h(b) \leq v^*(b)$), guarantees convergence to an optimal or bounded suboptimal solution. However, constructing effective admissible heuristics is often challenging. An admissible heuristic requires the search to explore all relevant portions of the belief space to guarantee that the optimal solution is not overlooked. This results in large convergence times. On the contrary, inadmissible heuristics tend to be greedy. They encourage the search to quickly converge to a suboptimal solution by exploring a narrow region of the belief space.

% In our setting, using only an admissible heuristic resulted in the search exploring too large a belief space. Consequently, within a limited time budget, the search does not explore enough relevant sections to output reasonable partial policies. 

In our setting, using only an admissible heuristic resulted in the search extensively exploring and not converging to reasonable partial policies within the limited time budget.
Conversely, using an inadmissible heuristic alone led to a very suboptimal solution quickly due to extremely limited exploration. To balance the exploratory and exploitative nature of these heuristics, we use a combination of both admissible and inadmissible heuristics. We maintain both values for a given belief state. When broader exploration is needed, we choose the best action based on the admissible value. For finer optimization, we use the inadmissible heuristic.

We experimented with different schedules for utilizing these heuristics and found that using the admissible heuristic in the initial part of the time budget and leaning on the inadmissible heuristic later proved ideal (Lines \ref{alg_line: ad, inad begin} - \ref{alg_line: ad, inad end}). The admissible heuristic promotes exploring different homotopies of the cost manifold, aiming for a global optimum. The inadmissible heuristic, in turn, helps find a local minimum within a homotopy. By exploring various homotopies initially and fine-tuning the solution with the greedy heuristic later, we achieve a balanced and effective search strategy.

\section{RESULTS} \label{section: Results}
The performance of the overall framework is demonstrated both in the real world and in simulation on the task of inserting a plug into a port using a UR10e manipulator under different magnitudes of pose uncertainties.

\subsection{Real World Robustness}

The planner's performance was evaluated in the real world under two different setups. In the first setup, the robot is tasked with localizing a plug on the wall of a shelf, with its exact pose unknown due to obstruction (Fig. \ref{fig:shelf_setup}). This involves in-plane localization ($SE(2)$), where the robot must resolve 3-dimensional pose uncertainty ($y, z, roll$), with positional uncertainty of 50 cm and angular uncertainty of $2\pi$. In the second, the port was mounted on a movable base, requiring the robot to resolve 4-dimensional pose uncertainty ($x, y, z, yaw$), with positional uncertainties of 40 cm and yaw uncertainties of 50 degrees. The framework was evaluated on 30 real-world runs for each setup, with the port placed in random poses within its workspace. It succeeded in 28 out of 30 runs for the shelf setup and 27 out of 30 runs for the movable base setup, with average execution times of 33.2 seconds and 42.3 seconds, respectively. The few failures occurred when our method eliminated all possible particles. This happened because our initial hypothesis set was created by sampling at a discretized resolution, and the true pose was not close enough to a sampled particle. Increasing the sampling resolution would mitigate this issue, albeit at the expense of computation.

\subsection{Simulation Comparisons}

\begin{table}[]
\begin{center}
\caption{Performance comparison of the proposed planner against baselines (all metrics presented are relative to the performance of the proposed planner)}
\label{ch4_tab: planner_comparison_baselines}
\begingroup
\setlength{\tabcolsep}{4pt}
\begin{tabular}{@{}l@{\hskip 0.15in}c@{\hskip 0.15in}c@{\hskip 0.15in}c@{\hskip 0.15in}c@{}}
\toprule
\multirow{2}{*}{\textbf{Planner}} & \multirow{2}{*}{\textbf{Cost}} & \textbf{Total plan} & \multirow{2}{*}{\textbf{Num Iters}}  & \textbf{Plan time} \\
& & \textbf{time} & & \textbf{per iter} \\
\midrule
\shortstack{Ours} & 1 & 1 & 1 & 1 \\ \midrule
\shortstack{TBL} & 2.12	& 0.72 & 1.36 & 0.51 \\ \midrule
\shortstack{Frontier} & 1.73 & 0.66 & 3.09 & 0.21 \\ \midrule
\end{tabular}\vspace{-0.3cm}
\endgroup
\end{center}
\end{table}

The framework's performance was evaluated in simulation on the plug insertion task in the presence of 5D pose uncertainty ($x, y, z, roll, yaw$). The positional uncertainty was on the order of 30 centimeters and the angular uncertainty was on the order of 30 degrees. 

% \begin{figure}
% \centering
%   \includegraphics[width=0.4\textwidth]{figs/num_particles_vs_time.eps} 
%   \caption{Number of particles vs Time required to evaluate 10 actions} \label{ch4_fig: num_particles_vs_time}
% \end{figure}

% The time required to evaluate 10 actions increases (roughly) linearly with the number of particles, with the evaluations taking 7 seconds for 5000 particles and over 60 seconds for 40000. This demonstrates the necessity of the hierarchical uncertainty representation, as evaluating even 10 actions becomes impractically slow when the number of particles exceeds 5000. Given the large number of particles typically involved (in the millions) and the need to evaluate thousands of actions per planning cycle, direct particle space reasoning is infeasible. It also must be reiterated that the proposed hierarchical representation is independent of the planner employed and makes planning under large magnitudes of pose uncertainty feasible.
The time required to evaluate actions increases linearly with the number of particles, with 10 evaluations taking over 60 seconds for 40000 particles. Given the large number of particles typically involved (in the millions) and the need to evaluate thousands of actions per planning cycle, direct particle space reasoning is infeasible. We also reiterate that the proposed hierarchical representation is independent of the planner employed and makes planning under large magnitudes of pose uncertainty feasible.

% The time required to evaluate 10 actions increases (roughly) linearly with the number of particles, with the evaluations taking over 60 seconds for 40000 particles. Direct particle space reasoning for large pose uncertainty requires reasoning about millions of particles with thousands of actions and is thus infeasible. Our proposed hierarchical representation makes planning under large pose uncertainty feasible.

\begin{table}[]
\begin{center}
\caption{Performance comparison of the proposed planner against baselines (all metrics presented are relative to the performance of the proposed planner)}
\label{ch4_tab: planner_comparison_baseline_ablation}
\begingroup
\setlength{\tabcolsep}{4pt}
\begin{tabular}{@{}l@{\hskip 0.15in}c@{\hskip 0.15in}c@{\hskip 0.15in}c@{\hskip 0.15in}c@{}}
\toprule
\multirow{2}{*}{\textbf{Planner}} & \multirow{2}{*}{\textbf{Cost}} & \textbf{Total plan} & \multirow{2}{*}{\textbf{Num Iters}}  & \textbf{Plan time} \\
& & \textbf{time} & & \textbf{per iter} \\
\midrule
& \multicolumn{4}{c}{\multirow{2}{*}{\textbf{Particle Space}}}\\\\ 
\cmidrule{2-5}
\shortstack{Ours} & 1 & 1 & 1 & 1 \\ \midrule
\shortstack{TBL} & 2.83 & 0.92 & 1.46 & 0.63\\ \midrule
\shortstack{Frontier} & 2.36 & 0.71 & 2.79 & 0.25 \\ \midrule
& \multicolumn{4}{c}{\multirow{2}{*}{\textbf{Volumetric Space}}}\\\\ 
\cmidrule{2-5}
\shortstack{Ours} & 1 & 1 & 1 & 1 \\ \midrule
\shortstack{TBL} & 1.94 & 0.63 & 1.34 & 0.47 \\ \midrule
\shortstack{Frontier} & 1.52 & 0.65 & 3.20 & 0.20 \\ \midrule
\end{tabular}\vspace{-0.3cm}
\endgroup
\end{center} \vspace{-1em}
\end{table}

We also compare the performance of our planning framework against two greedy baselines, TBL and Frontier (Table \ref{ch4_tab: planner_comparison_baselines}). TBL is a popular online planning framework that interleaves planning and execution. In each iteration of planning, TBL samples multiple actions and executes the one that maximizes an information gain metric. Based on the observation, the belief of the system state is updated and the process is repeated. On the other hand, Frontier search identifies the nearest action that provides information gain and executes it in the real world. The 1-step greediness of both approaches is reflected in their accumulated cost (distance traveled by the robot), which our framework reduces by more than 50\%. 

For each iteration of planning, a time budget of 1 second was provided for all of the planners. As frontier search returns a solution immediately after identifying the nearest information gain action, it takes the least time per iteration. We observe that although both of the baselines take less time per iteration, they need significantly more number of iterations to complete the task. However, their total time continues to be on the lower side in comparison to our framework. We also present ablations of the planners' performances in the particle space and the volumetric space (Table \ref{ch4_tab: planner_comparison_baseline_ablation}). Our framework consistently outperforms the baselines in terms of cost incurred, achieving up to 2.8 times lower cost in the particle space and 1.9 times lower cost in the volumetric space. Given that a large portion of the problem is solved in the volumetric space, our framework demonstrates a net cost improvement of up to 2.12 times.

\begin{table}[]
\begin{center}
\caption{Evaluating the impact of combining admissible and inadmissible heuristics (all metrics presented are relative to the performance of the combined planner)}
\label{ch4_tab: planner_ablation_heur_combination}
\begingroup
\setlength{\tabcolsep}{4pt}
\begin{tabular}{@{}l@{\hskip 0.15in}c@{\hskip 0.15in}c@{\hskip 0.15in}c@{\hskip 0.15in}c@{}}
\toprule
\multirow{2}{*}{\textbf{Planner}} & \multirow{2}{*}{\textbf{Cost}} & \textbf{Total plan} & \multirow{2}{*}{\textbf{Num Iters}}  & \textbf{Plan time} \\
& & \textbf{time} & & \textbf{per iter} \\
\midrule
\shortstack{Ours} & 1 & 1 & 1 & 1 \\ \midrule
\shortstack{RTDP-Inad} & 1.29 & 0.88 & 1.15 & 0.76 \\ \midrule
\end{tabular}\vspace{-0.3cm}
\endgroup
\end{center}
\vspace{-1em}
\end{table}

Table \ref{ch4_tab: planner_ablation_heur_combination} illustrates the impact of combining admissible and inadmissible heuristics. Using only an inadmissible heuristic causes the search to quickly converge to a suboptimal solutions. This is reflected in the higher costs and shorter iteration times. In contrast, combining it with an admissible heuristic improves solution quality by up to 30\% by enabling exploration of a larger portion of the space. Solely using an admissible heuristic results in excessive exploration of the belief space, failing to identify effective policies within the time budget and completing the task. Hence, its results are not included in the table.

\section{CONCLUSION}
In this work, we present an online planning framework for unstructured settings, enabling robots to use binary contact signals to reduce large pose uncertainties and complete manipulation tasks. To manage the computational complexity, we employ a hierarchical approach. Initially, we reason in a coarse 3D volumetric space; once uncertainty is sufficiently reduced, we transition to particle space for finer adjustments. This method significantly reduces planning complexity, enabling real-time problem solving under large uncertainties. Our closed-loop framework uses a heuristic search-based anytime solver to compute partial policies within a limited time budget. Our approach achieved a 93\% real-world success rate and improved solution quality by over 50\% compared to greedy baselines on a high-precision task.

\bibliographystyle{plain}
\bibliography{references}

\end{document}